\documentclass[10pt,twocolumn,letterpaper]{article}

\usepackage{iccv}
\usepackage{times}
\usepackage{epsfig}
\usepackage{graphicx}
\usepackage{amsmath}
\usepackage{amssymb}

\usepackage[utf8]{inputenc}
\usepackage{amssymb}
\usepackage{amsmath}
\usepackage{amsfonts}   
\usepackage{dsfont}
\usepackage{algorithm}
\usepackage{graphicx}
\usepackage[T1]{fontenc}       
\usepackage{url}            
\usepackage{booktabs}       
\usepackage{nicefrac}       
\usepackage{microtype}      
\usepackage{lipsum}
\usepackage{algpseudocode}
\usepackage{xcolor}
\usepackage{multirow}
\usepackage{comment}
\usepackage{booktabs}

\usepackage[breaklinks=true,bookmarks=false]{hyperref}

\iccvfinalcopy 

\def\iccvPaperID{11969} 

\ificcvfinal\fi

\begin{document}

\title{Multimodal Parameter-Efficient Few-Shot Class Incremental Learning}

\author{Marco D'Alessandro$^{1*}$ \hspace{0.5cm} Alberto Alonso$^{1*}$ 
\\
Enrique Calabrés$^{1,2}$ \hspace{0.5cm} Mikel Galar$^{2}$
\\
\\
$^1$Neuraptic AI \hspace{0.5cm} $^2$Public University of Navarra
\\
{\tt\small \{marco.dalessandro, alberto.alonso, enrique.hernandez\}@neuraptic.ai}
\\
{\tt\small mikel.galar@unavarra.es}
}

\maketitle
\ificcvfinal\thispagestyle{empty}\fi

\begin{abstract}
Few-Shot Class Incremental Learning (FSCIL) is a challenging continual learning task, where limited training examples are available during several learning sessions. To succeed in this task, it is necessary to avoid over-fitting new classes caused by biased distributions in the few-shot training sets. The general approach to address this issue involves enhancing the representational capability of a pre-defined backbone architecture by adding special modules for backward compatibility with older classes. However, this approach has not yet solved the dilemma of ensuring high classification accuracy over time while reducing the gap between the performance obtained on larger training sets and the smaller ones. In this work, we propose an alternative approach called Continual Parameter-Efficient CLIP (CPE-CLIP) to reduce the loss of information between different learning sessions. Instead of adapting additional modules to address information loss, we leverage the vast knowledge acquired by CLIP in large-scale pre-training and its effectiveness in generalizing to new concepts. Our approach is multimodal and parameter-efficient, relying on learnable prompts for both the language and vision encoders to enable transfer learning across sessions. We also introduce prompt regularization to improve performance and prevent forgetting. Our experimental results demonstrate that CPE-CLIP significantly improves FSCIL performance compared to state-of-the-art proposals while also drastically reducing the number of learnable parameters and training costs.
\let\thefootnote\relax\footnotetext{$^*$Equal contribution.}
\end{abstract}

\section{Introduction}

Deploying ML systems in a dynamic environment requires accounting for continuous data streams arriving over time. This environment may experience shifts in data distribution or the addition of new classes. An ideal learning system must be able to learn new incoming classes while maintaining its discriminability over previously learned classes, thus avoiding catastrophic forgetting \cite{mccloskey1989catastrophic}. This continual learning problem formulation is known as Class-Incremental Learning (CIL), which requires dealing with the stability-plasticity dilemma \cite{mermillod2013stability, hebb2005organization}, \ie, the trade-off between learning new classes and retaining old ones. In this work, we focus on a special case of CIL, named Few-Shot Class Incremental Learning (FSCIL, \cite{tao2020few}), where only a few training examples are available at every learning session. Here, the additional challenge consists in avoiding over-fitting on new incoming classes caused by biased distributions in the few-shot training sets. This problem is particularly crucial in practical, real-world scenarios where data availability is limited. Examples of such scenarios include manufacturing settings \cite{yu2018one, ayub2020tell} and medical imaging \cite{jiang2021few}. In manufacturing, robots are deployed to carry out a diverse range of tasks, such as assembling or grasping objects. To perform these tasks, robots may need to adapt to new objects or materials, which may have a limited amount of training data available. In medical imaging the availability of data may also be limited due to the high costs of data collection and patient privacy, making it difficult to acquire new knowledge over time.

Recent research has focused on solving these problems through various approaches, such as meta-learning \cite{yin2022sylph, przewikezlikowski2022hypermaml}, regularization techniques \cite{mazumder2021few}, or knowledge distillation \cite{rebuffi2017icarl, cheraghian2021semantic, zhou2022forward}. These methods have shown promising results in achieving incremental learning over time with a limited amount of data available. The general approaches consist in enhancing the basic representational capability of a predefined backbone architecture by adding special modules to entail backward compatibility with older classes during learning sessions. These solutions are computationally expensive since they need a large number of iterations in each session to adapt the additional modules to new classes while maintaining backward compatibility. Despite the high computational cost, they still fail to efficiently reduce the gap between the performance obtained on larger training sets and the one obtained on smaller sets over time, which still remains an unsolved dilemma \cite{zou2022margin, liu2020negative}. 

In this work, we propose Continual Parameter-Efficient CLIP (CPE-CLIP) as an alternative approach to reduce the loss of information between different learning sessions. Inspired by the astounding continual learning performance obtained in a zero-shot setting \cite{thengane2022clip}, we use CLIP \cite{radford2021learning} as a starting point to build a continual learning system for FSCIL. Instead of relying on adapting additional modules to address information loss, we propose to adapt the CLIP architecture with lightweight learnable prompts for few-show image classification. In this way, we are able to take advantage of the vast amount of knowledge acquired by CLIP in large-scale pretraining and its inherent effectiveness in generalizing to new concepts. Notably, this is a Multimodal and Parameter-Efficient approach as it relies on learnable prompts, rather than finetunig, of both the language and vision encoders to allow transfer learning across sessions over time. We show that our approach significantly improves FSCIL performance compared to state-of-the-art proposals while also drastically reducing the number of learnable parameters and training costs. We also conduct extensive hyperparameter tuning and ablation studies to understand the functional properties of the multiple components of our model. Our main contributions can be summarized as follows:
\begin{itemize}
    \item We propose a prompt-based approach to adapt the CLIP architecture for solving continual learning tasks in few-shot settings by reducing forgetting and supporting knowledge transfer over time, all while learning less than $0.3\%$ of the total parameters.
    \item We combine two different prompt attachment methods with prompt regularization to smoothly transition to future tasks while maintaining constant performance over time.
    \item We achieve state-of-the-art performance on three popular benchmark datasets for FSCIL, and exceed previous state-of-the-art results by a great margin.
\end{itemize}

The rest of this paper is organized as follows. We discuss related works about our methodological approach and few-shot class-incremental learning in Section 2. Section 3 introduces the problem formulation. The proposed method is presented in Section 4. Moreover, we present our experimental results and final considerations in Sections 5 and 6, respectively.

\section{Related Work}

Our approach to the few-shot continual learning problem is related to several topics, so we introduce them separately.

\subsection{Few-Shot Image Classification} 

Few-shot image classification aims to fit new unseen classes with an insufficient number of training examples \cite{chen2019closer, wang2020generalizing}. Several learning methods have been proposed for this purpose. For instance, in metric-based approaches, different network branches are built to classify images by calculating the distance between a query image from the test set and the training images in the few-shot training set \cite{snell2017prototypical, vinyals2016matching, sung2018learning}. Differently, in meta-learning, models are trained on a variety of learning tasks, such that they can solve new learning tasks using only a small number of training samples \cite{finn2017model, elsken2020meta, yin2022sylph, przewikezlikowski2022hypermaml}. A rather different and more recent perspective relies on pretrained multimodal vision-language models to classify images from labeled captions with minimal training examples \cite{zhou2022learning, zhou2022conditional, khattak2022maple}. In this approach, few-shot learning is based on the correct match between the query image and a text caption describing the category label. Our method can be seen as a continual learning adaptation of the latter approach.

\subsection{Incremental Learning}

Incremental learning deals with the problem of learning new information from non-stationary data streams \cite{van2022three, mai2022online, qu2021recent}. According to the availability of task identifiers (IDs) over time, the problem formulation can pertain to either task- or class-incremental learning. There are several solutions in the literature that try to face these tasks by enabling learning of incoming information from new tasks while minimizing forgetting of previously acquired knowledge. In regularization-based methods, specific parameters are regularized for learned tasks in order to retain knowledge acquired on previous ones and avoid catastrophic forgetting \cite{kirkpatrick2017overcoming, li2017learning, zenke2017continual}. Architecture-based methods assign an isolated portion of the backbone, or isolated parameters of additional branches, to each task \cite{rusu2016progressive, yoon2017lifelong, mallya2018packnet, wang2020learn, ebrahimi2020adversarial}. In rehearsal-based methods, data from previously learned tasks are stored in a rehearsal buffer and used in the current task in addition to the current training set \cite{buzzega2020dark, cha2021co2l, pham2021dualnet}. A more recent prompt-based rehearsal-free approach combines powerful pretrained backbones with learnable prompts that retain the knowledge acquired from the different tasks without modifying the weights of the main backbone, thus avoiding forgetting \cite{wang2022dualprompt, wang2022learning, smith2022coda, razdaibiedina2023progressive}. Our method gets inspiration from the solutions proposed in the latter methods.

\subsection{Few-Shot Class-Incremental Learning} 

FSCIL is a recent research topic proposed to tackle few-shots training inputs in a class-incremental setting \cite{tao2020few}, where task ID is not provided during evaluation. The general task is to initially learn from a number of base classes and then continuously update the model to represent new incoming classes. The main challenge of this setting is to avoid overfitting to new class few-shot samples. The first attempt to solve this issue proposed the \textit{neural gas} (NG) network for representing knowledge \cite{tao2020few}, where feature space topologies were learned for different classes, and new classes were represented by growing and adapting the network's topology. In \cite{zhang2021few}, authors decoupled learning representations and classifiers, by letting the latter be updated over time by means of a graph model propagating information between classifiers. Prototype modeling was also used to assign prototypes in the embedding space to reserve it for future incoming classes \cite{zhou2022forward}, or to use the average of new class embedding representations as a class prototype to replace classifiers \cite{zhou2022few}. Different methods addressed the problem by synthesizing features into a mixture of sub-spaces for incremental classes by using a VAE \cite{cheraghian2021synthesized}, or by adapting general deep learning architectures to enable a few parameters to be updated for every new set of novel incoming classes \cite{mazumder2021few}. More recent approaches tried to combine features emerging from supervised and self-supervised models for boosting classifiers \cite{ahmad2022few}, or to calibrate distributions to avoid forgetting by retrieving distributions for old classes while estimating distributions for new classes \cite{liu2022learnable}.

\subsection{Parameter-Efficient Learning}

The most common way to adapt foundational large general-purpose pretrained models to downstream tasks is to finetune all the model parameters, which results in high computational costs and memory usage, and the need to store several copies of the finetuned model for different tasks. A lightweight alternative came from the parameter-efficient learning literature that proposed to update only a small number of extra parameters while keeping backbone parameters frozen \cite{he2021towards, lester2021power}. Several methods have been proposed to flexibly adapt pretrained backbones to different downstream tasks according to this logic. Adapter-tuning \cite{houlsby2019parameter, houlsby2020k} interleaves transformer layers with a feed-forward bottleneck module with skip-connection to adapt the layer's output before passing to the next layer. Prefix-tuning \cite{li2021prefix, wang2022dualprompt, jia_visual_2022} prepends tunable prefix vectors as learnable embeddings to the keys and values of the multi-head attention layers in transformers \cite{vaswani2017attention}. In prompt-tuning \cite{lester2021power, wang2022learning}, a set of learnable embeddings is prepended to the input embeddings from the first layer, and the augmented input is then normally processed by the frozen transformer layers.

\section{Problem Formulation} 

The FSCIL setting \cite{tao2020few} can be defined as follows. We consider a stream of labelled training sets $D_0, D_1, \ldots, D_T$, where $ D_t = \{ (\boldsymbol{x}_{i,t}, y_{i,t}) \}_{i=1}^{N^D_t} $, $N^D_t$ is the number of training examples provided at session $t$, and $T$ is the last incremental session. $D_0$ identifies the large-scale training set of base classes, and $D_t$ is the few-shot training set of new classes, for $t>0$. The base class dataset, $D_0$, is meant to have a sufficient number of training examples. On the contrary, insufficient training sets are provided for new classes. Consider the set of class labels $C_t$ belonging to train set $D_t$. FSCIL has the following requirements: (1) classes don't overlap among sessions, $\forall t,\tau$, $ t \neq \tau$, $C_t \cap C_{\tau} = \varnothing$, (2) base class set is bigger than new class sets, where $|C_0| > |C_t|$, and $N^D_0 > N^D_t$, hold for $t > 0$, (3) new class sets have the same size, such that $\forall t,\tau$, $ t \neq \tau$, $|C_t| = |C_{\tau}|$ and $N^D_t = N^D_{\tau}$ hold for $t,\tau > 0$. For the evaluation phase, the only requirement is that session-wise performances are computed by considering all the classes encountered up to the current session $t$. Consider the stream of labelled evaluation sets $E_0, E_1, \ldots, E_T$, and a model $f$, than the evaluation accuracy for session $t$ can be computed as follows:

\begin{align}
    A_t = \frac{\displaystyle \sum_{(\boldsymbol{x}_i, y_i) \in E_0 \cup E_1 \cup \ldots E_t} \left[ f(\boldsymbol{x}_i) = y_i \right]}{N_0^E + N_1^E + \ldots + N_t^E}
\end{align}
\\
where $N^E_{\tau}$ is the number of evaluation examples for session $\tau$, and $[\cdot]$ the indicator function.

\section{Method}

\begin{figure*}
    \begin{center}
        \includegraphics[width=0.9\linewidth]{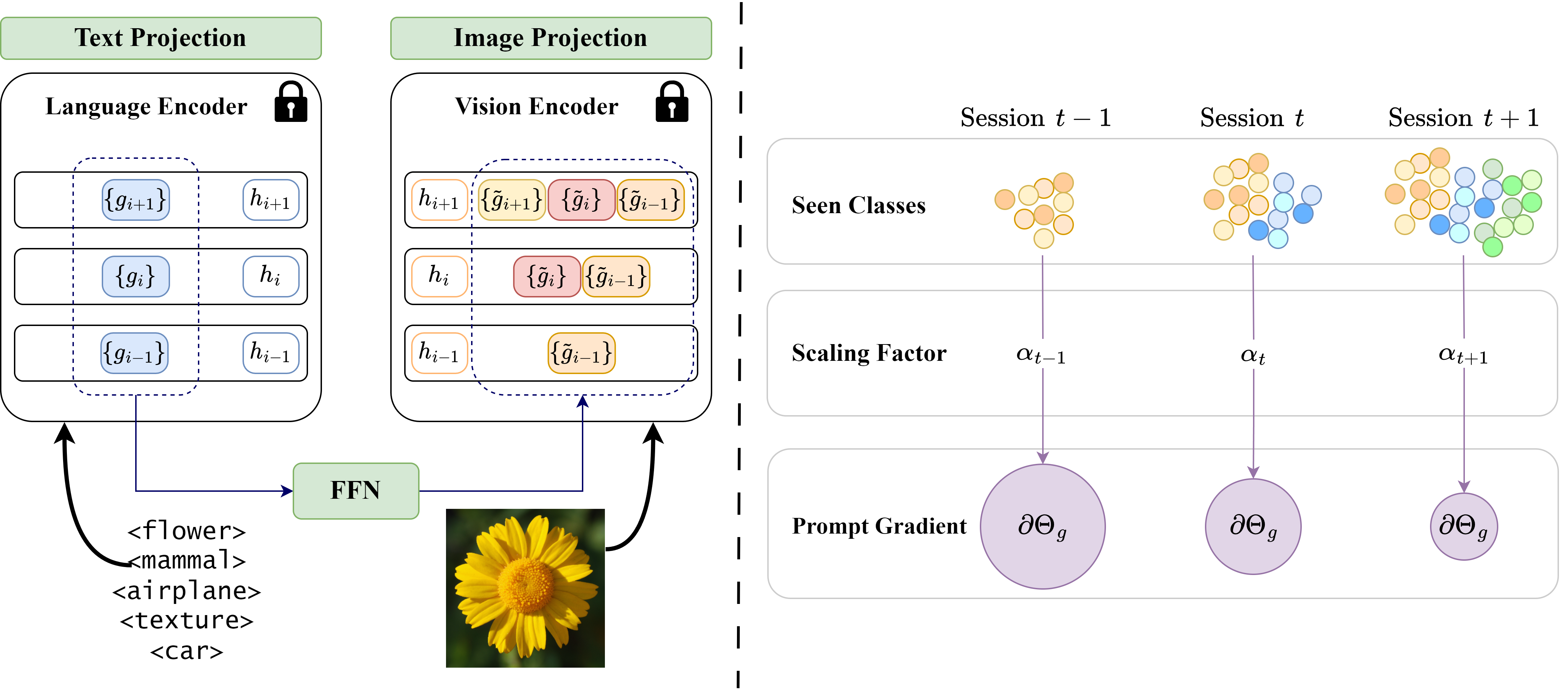}
    \end{center}
    \caption{Summary of CPE-CLIP architecture and training process. The picture on the left describes the general structure of CPE-CLIP, where the G-Prompt contributes to generalizing task-invariant knowledge on the language encoder and is then projected to the vision encoder. Vision prompts are accumulated across subsequent layers, while a replacement strategy is used for the language encoder. The image on the right depicts the regularization process where an increase in the number of seen classes reduces parameter gradients by means of the scaling factor $\alpha_t$, for a given session $t$.}
    \label{fig:Fig.1}
\end{figure*}

Our proposed method, CPE-CLIP, is summarized in Figure \ref{fig:Fig.1}. The approach involves the use of the CLIP foundational vision-language model \cite{radford2021learning} as the primary building block of a continual learning system. CLIP is a neural network trained to align the modalities of language and vision, and to leverage the abundant supervision offered by natural language to reason about visual concepts. In our work, we exploit the capabilities of CLIP to cast image classification as a multimodal task, where text prompts (\eg "a photo of a <category>") are used as query captions for the text encoder, and the matching of a test image to the caption serves as the classification criterion. Our approach learns prompts that are adapted to both the language and vision encoders \cite{khattak2022maple}. By doing so, we maintain the knowledge acquired during pretraining by freezing the CLIP backbone while allowing the prompts to solve the continual learning task. To further enhance performance and avoid forgetting, we introduce prompt regularization.

\textbf{Language Encoder.} Here we learn language context prompts that are shared among all the classes of a given downstream task. Context prompts fulfill two purposes: (1) they prevent manually selecting inefficient prompts \cite{zhou2022learning, zhou2022conditional} to provide the textual representation of an image, assuming the category label of that image is available, (2) they serve as stability parameters that learn \textit{general} task-invariant properties shared among the session tasks. We then introduce $L$ learnable tokens, $\boldsymbol{g}$, called G-Prompt (by following notation in Wang \etal \cite{wang2022dualprompt}), such that $\boldsymbol{\Theta}_{\boldsymbol{g}} \in \mathbb{R}^{L \times d^{\textrm{NLP}}}$, where $d^{\textrm{NLP}}$ is the embedding dimension of CLIP language encoder. The input embeddings now follow the form $[g_1, g_2, \ldots, g_L, w]$ = $[\boldsymbol{g}, w]$, and $w$ is the embedding for the category name of the input image. Let's define $f^{\textrm{NLP}}_i$ as the $i^{th}$ transformer layer of the language encoder, where $i=1,2,\ldots, K$ and $K$ is the total number of layers. We admit the case where new learnable tokens can be further introduced in each language encoder layer, $f^{\textrm{NLP}}_i$, up to a certain depth $D$. In this way, different prompts are used independently across different layers to account for different levels of abstraction. The forward pass can then be described as:

\begin{equation}
    [\_, h_1] = f^{\textrm{NLP}}_1([\boldsymbol{g}_1, w]) 
    \label{eq.2}
\end{equation}
\\
 for the first layer, and 
 
\begin{equation}
    [\_, h_i] = f^{\textrm{NLP}}_i([\boldsymbol{g}_{i-1}, h_{i-1}]) 
    \hspace{1cm}
    i=2,3,4, \ldots, D
    \label{eq.3}
\end{equation}
\\
for subsequent layers up to layer $D$. Here, $[\cdot,\cdot]$ refers to vertical concatenation, $h_i$ is the latent state output from layer $i$ associated with the class category word embedding, and $\boldsymbol{g}_i$ represents the set of learnable prompts for layer $i$. In this phase, the output embeddings for the input prompt $\boldsymbol{g}_i$ are discarded for the next layer. Notably, when layer-specific prompts are introduced we have  $\boldsymbol{\Theta}_{\boldsymbol{g}} \in \mathbb{R}^{D \times L \times d^{\textrm{NLP}}}$. After $D^{th}$ transformer layer, subsequent layers process previous output layers in a standard way until the final text representation:

\begin{equation}
    [\boldsymbol{g}_i, h_i] = f^{\textrm{NLP}}_i([\boldsymbol{g}_{i-1}, h_{i-1}])
    \hspace{0.7cm}
    i=D+1, \ldots, K.
    \label{eq.4}
\end{equation}
\\
For simplicity, we refer to the last hidden state for the [EOS] token as $h^{\textrm{NLP}}$, which in the CLIP language encoder is used to represent the whole sentence. The hidden state is then projected to
a lower dimensional space:

\begin{equation}
    h_*^{NLP} = p^{\textrm{NLP}}(h^{\textrm{NLP}}),
    \label{eq.5}
\end{equation}
\\
where $p^{\textrm{NLP}}$ is a linear projection layer, and $h_*^{\textrm{NLP}}$ is the final low-dimensional vector for the text representation.

\textbf{Vision Encoder.} As for the language encoder, we conceive G-Prompt for the vision branch. Even in this case, prompts are concatenated with image patch embeddings across several layers of the hierarchy in order to interact with lower and higher-level image feature processing. We introduce $L$ tokens, $\boldsymbol{\Tilde{g}}$, such that $\boldsymbol{\Tilde{g}} \in \mathbb{R}^{L \times d^{\textrm{CV}}}$, where $d^{\textrm{CV}}$ is the embedding dimension of CLIP vision encoder, and $d^{\textrm{CV}} > d^{\textrm{NLP}}$. The input embeddings now follow the form $[c_1, c_2, \ldots, c_J, \Tilde{g}_1, \Tilde{g}_2, \ldots, \Tilde{g}_{L}]$ = $[\boldsymbol{c}, \boldsymbol{\Tilde{g}}]$, where $\boldsymbol{c}$ is the embedded patches set of the input image plus the additional [CLS] token, and $J$ the total number of embeddings. Let's now define $f_i^{\textrm{CV}}$ as the $i^{th}$ transformer layer of the vision encoder, where $i=1,2,\ldots, K$. Similar to  Khattak \etal \cite{khattak2022maple}, we bridge the gap between language and vision prompts by explicitly expressing the latter as a function of the former. We introduce a learnable linear projection $f^{\textrm{PROJ}}$, $\boldsymbol{\Theta}_{f^{\textrm{PROJ}}} \in \mathbb{R}^{d^{\textrm{NLP}} \times d^{\textrm{CV}}}$, and constraint vision task-invariant prompts to be conditioned on language G-Prompt, such that $\boldsymbol{\Tilde{g}}_i = f^{\textrm{PROJ}}(\boldsymbol{g}_i)$, and $\boldsymbol{\Tilde{g}}_i$ is the set of prompts for vision encoder layer $i$. Although prompts in the vision branch are derived from language context prompts, we found it beneficial to use a different strategy for propagating prompts across the layers hierarchy. Here, we propose an \textit{accumulation} method, as an alternative to the \textit{replacement} method used in the language branch, where prompts in different layers are not independent anymore since they can interact with the processed output embeddings of prompts from previous layers. As usual, prompt accumulation takes place up to depth $D$. Formally, we describe the forward pass in the vision encoder as:

\begin{equation}
    [h_1, \boldsymbol{\Tilde{g}}_1] = f^{\textrm{CV}}_1([\boldsymbol{c}, \boldsymbol{\Tilde{g}}_1])
    \label{eq.6}
\end{equation}

\noindent for the first layer, and

\begin{equation}
    \begin{split}
        \boldsymbol{\Tilde{g}}_i &= [\boldsymbol{\Tilde{g}}_i, \boldsymbol{\Tilde{g}}_{i-1}] \\
        [h_i, \boldsymbol{\Tilde{g}}_i] &= f^{\textrm{CV}}_i([\boldsymbol{c}, \boldsymbol{\Tilde{g}}_i])
    \end{split} 
    \hspace{1cm}
    i=2,3,4, \ldots, D
    \label{eq.7}
\end{equation}
\\
for subsequent layers up to layer $D$. After $D^{th}$ transformer layer, prompts are not accumulated anymore, and subsequent layers process previous output layers in a standard way until the final image representation:

\begin{equation}
    [\boldsymbol{\Tilde{g}}_i, h_i] = f^{\textrm{CV}}_i([\boldsymbol{\Tilde{g}}_{i-1}, h_{i-1}])
    \hspace{1cm}
    i=D+1, \ldots, K
\end{equation}
\\
where now $\boldsymbol{\Tilde{g}}_i$ is the final pooled prompt such that $\boldsymbol{\Tilde{g}}_i \in \mathbb{R}^{LD \times d^{\textrm{CV}}}$. We refer to the last hidden state related to the [CLS] token as $h^{\textrm{CV}}$, which in the CLIP vision encoder is used to represent the whole image. The hidden state is then projected to a lower dimensional space:

\begin{equation}
    h_*^{\textrm{CV}} = p^{\textrm{CV}}(h^{\textrm{CV}})
\end{equation}
\\
where $p^{\textrm{CV}}$ is a linear projection layer, and $h_*^{\textrm{CV}}$ is the final low-dimensional vector for the image representation.

\textbf{Multimodal Classification.} The prediction probability for every given input image $x$ to be classified as belonging to class $z$, $z=1,2,\ldots,Z$, is computed as:

\begin{equation}
    p(y=z|x) = \frac{\exp[ \rho(h_*^{\textrm{CV}}, h_{*z}^{\textrm{NLP}}) ]}{\textrm{\Large{$\Sigma$}}_{j=1}^Z  \exp[ \rho(h_*^{\textrm{CV}}, h_{*j}^{\textrm{NLP}}) ]}
\end{equation}
\\
where $\rho$ is the \textit{cosine similarity}, $h_*^{\textrm{CV}}$ the projected representation of image $x$, and $h_{*z}^{\textrm{NLP}}$ is the projected representation of the sentence with the category name of $z^{th}$ class in the training set.

\textbf{Prompt Regularization.} In FSCIL, base class training is crucial to initially tune the network to boost generalization to novel classes. In our case, the G-Prompt introduces a set of tokens to fulfill this purpose. Such tokens provide an efficient text representation that can be matched with an image in order to correctly classify the latter as belonging to the correct (semantic) category/label. However, the base class set provides a greater chance for generalization compared to session-related class sets, since it provides a greater number of classes and training examples. For this reason, we propose a mechanism to preserve knowledge proportionally to the number of classes encountered in different sessions. We define a scaling factor $\alpha_t$ for a given session $t$, $t=1,2,\ldots, T$, that affects the updating rate of G-Prompt parameters when training on session $t$:

\begin{equation}
    \alpha_t = \frac{|C_t|}{ \sum_{\tau=0}^t |C_\tau|}.
\end{equation}
\\
We then apply regularization as follows:

\begin{equation}
    \frac{\partial \mathcal{L}_t}{\partial \boldsymbol{\Theta}_{\boldsymbol{g}}} \gets \alpha_t \frac{\partial \mathcal{L}_t}{\partial \boldsymbol{\Theta}_{\boldsymbol{g}}}
\end{equation}

\noindent and

\begin{equation}
    \frac{\partial \mathcal{L}_t}{\partial \boldsymbol{\Theta}_{f^{\textrm{PROJ}}}} \gets \alpha_t \frac{\partial \mathcal{L}_t}{\partial \boldsymbol{\Theta}_{f^{\textrm{PROJ}}}}
\end{equation}
\\
where $\mathcal{L}_t$ is the loss function for the classification task in session $t$. Such a regularization allows the G-Prompt, as well as the language-vision prompt projection, to be updated less consistently as the number of total seen classes increases.

\section{Experiment}

We evaluate CPE-CLIP on the three benchmarks \cite{tao2020few} that provide the main baseline for model comparison in the FSCIL literature. Benchmarks include CIFAR100 \cite{krizhevsky2009learning}, \textit{mini}ImageNet \cite{russakovsky2015imagenet}, and CUB200-2011 \cite{wah2011caltech}.

\subsection{Evaluation Benchmarks} 
For all the benchmarks we follow the split proposed by Tao \etal \cite{tao2020few} since they provide the standard for all the proposals in the literature and ensure a fair model comparison. Benchmarks are described as follows:
\begin{itemize}
    \item \textbf{CIFAR100} The dataset contains 60.000 $32\times 32$ RGB images from 100 classes. We use 60 classes as the base class set. The remaining 40 classes are split into 8 sessions where each session contains 5 new classes, and the few-shot training set consists of 5 examples per class (5-way 5-shot incremental task).
    \item \textbf{\textit{mini}ImageNet} The dataset contains 60.000 $84\times 84$ RGB images. We use 100 classes as the base class set. The remaining 40 classes are split into 8 sessions of 5 few-shot training examples each (5-way 5-shot incremental task).
    \item \textbf{Caltech-UCSD Birds-200-2011 (CUB200)} The dataset contains 11.788 finegrained $224 \times 224$ RGB images from 200 classes of bird species. We use 100 classes as the base class set. The remaining 100 classes are partitioned into 10 sessions, or timestamps, where each session contains 10 new classes, and the few-shot training set consists of 5 examples per class (10-way 5-shot incremental task).
\end{itemize}

\subsection{Implementation Details}

We use the $16 \times 16$ patches OpenAI CLIP \cite{radford2021learning} version from the HuggingFace's Transformers library \cite{wolf-etal-2020-transformers} as the starting backbone. Models and pipelines are built in PyTorch with the aid of Avalanche library \cite{lomonaco2021avalanche}. We use the SGD optimizer with momentum, by setting learning rate to $0.00325$, weight decay to $1e^{-5}$, and a cosine annealing with warmup, for all the benchmarks. For the base class training we set batch size to $32$ and number of epochs to $3$ for CIFAR100 and \textit{mini}ImageNet, and batch size to $4$ with $6$ epochs for CUB200. For the new class session training sets we set batch size to $4$ and number of epochs to $5$ for all the benchmarks. All the experiments have been deployed on one single GeForce RTX 2080 Ti. For model comparison, we report the top-1 evaluation accuracy for base class and for every session since it is the standard practice in FSCIL. We also report the dropping rate (PD) metric, which measures the drop in accuracy in the last session \wrt the accuracy in base class session as a measure of forgetting, and the across-session average accuracy as a measure of overall performance.

\subsection{Hyperparameter Tuning}

We performed a hyperparameter tuning to select the best candidate model by varying the two hyperparameters affecting CPE-CLIP overall behavior on FSCIL. We used grid search to explore the following range of values: $L=[2,4]$, $D=[1,3,6,9,12]$ in an exhaustive $2\times5$ search. For every configuration, we use the average across-session accuracy of 5 runs with random parameter initialization as the metric for model selection. Due to the computational burden of the exhaustive hyperparameter search, we focused on CUB200 benchmark only, since it conveys a special challenge for our CLIP-based method due to the fine-grained images associated with technical, non-common, text labels reflecting bird species. Results are shown in Table \ref{table:tab1}

\begin{table}[H]
    \begin{center}
        \setlength{\tabcolsep}{4pt}
        \renewcommand{\arraystretch}{1.2}
        \begin{tabular}{cccccc}
        \toprule[1.5pt]
            & \multicolumn{5}{c}{$D$} \\
            \cmidrule(l){2-6}
            $L$ & 1 & 3 & 6 & 9 & 12 \\
            \hline
            \vspace{-2mm}
            2   & 67.48 & 68.63 & 68.54 & 69.49 & \textbf{70.23} \\
                & (0.25) & (0.38) & (0.38) & (0.14) & (0.31) \\
            \vspace{-2mm}
            4   & 68.03 & 68.35 & 69.61 & 69.11 & 69.28 \\
                & (0.26) & (0.07) & (0.45) & (0.36) & (0.28) \\
        \bottomrule[1.5pt]
        \end{tabular}
    \end{center}
    \caption{Hyperparameter Tuning for $L$ and $D$. Average across-session accuracy and standard errors (in brackets) for every configuration are reported.}
    \label{table:tab1}
\end{table}

The best model results in the hyperparameter set $L=2$, $D=12$. Therefore, we relied on this configuration for model comparison.

\subsection{Comparison with state-of-the-art models}

In this section, we show our main results on CIFAR100, \textit{mini}ImageNet, and CUB200 benchmarks, shown in Table \ref{table:tab2}, \ref{table:tab3}, and \ref{table:tab4}, respectively, where CPE-CLIP is compared with the latest state-of-the-art FSCIL approaches \cite{zhu2021self, ahmad2022few, zhang2021few, xu2023multi, xu2023flexible, zou2022margin, zhou2022forward, zhou2022few}. Models which are outperformed by a great margin by the most recent state-of-the-art methods were not included in the model comparison study.

According to these results, our model outperforms state-of-the-art models by a great margin. CPE-CLIP obtains the best classification accuracy in the base class session and reduces forgetting more efficiently, as shown by the PD metric, while maintaining stable high classification performances over time. CPE-CLIP shows superior abilities in reducing information loss when moving from training on a larger dataset, such as the base class set, to smaller datasets during learning sessions. It is worth mentioning that the other approaches included in the current model comparison primarily relied on ResNet \cite{he2016deep} and ViT \cite{dosovitskiy2020image} as the main backbones, which were pretrained solely for the CUB200 benchmark. Furthermore, CPE-CLIP relies on CLIP which was not pretrained directly on popular foundational datasets such as ImageNet \cite{russakovsky2015imagenet}, differently from ResNet and ViT.

\begin{table*}
    \begin{center}
        \setlength{\tabcolsep}{3pt}
        \renewcommand{\arraystretch}{1.1}
        \begin{tabular}{lccccccccc|cc|cc}
            \toprule[1.5pt]
            \multirow{2}{*}{Method} & \multicolumn{9}{c|}{Accuracy in each session (\%)} & \multirow{2}{*}{Avg. $\uparrow$} & \multirow{2}{*}{PD $\downarrow$} & \multirow{2}{*}{$\Delta$ Avg.} & \multirow{2}{*}{$\Delta$ PD} \\
            \cline{2-10}
            & 0 & 1 & 2 & 3 & 4 & 5 & 6 & 7 & 8 & & \\
            \hline
            CEC $\dagger$ \cite{zhang2021few} & 73.07 & 68.88 & 65.26 & 61.19 & 58.09 & 55.57 & 53.22 & 51.34 & 49.14 & 59.53 & 23.93 & \textbf{$+$23.89} & \textbf{$+$16.62} \\
            SPPR $\ddagger$ \cite{zhu2021self} & 76.33 & 72.33 & 67.33 & 63.33 & 59.00 & 55.33 & 53.00 & 50.33 & 47.33 & 60.47 & 29.00 & \textbf{$+$22.95} & \textbf{$+$21.69} \\
            CLOM $\dagger$ \cite{zou2022margin} & 74.2 & 69.83 & 66.17 & 62.39 & 59.26 & 56.48 & 54.36 & 52.16 & 50.25 & 60.57 & 23.95 & \textbf{$+$22.85} & \textbf{$+$16.64} \\
            FeSSSS $\dagger$ \cite{ahmad2022few} & 75.35 & 70.81 & 66.70 & 62.73 & 59.62 & 56.45 & 54.33 & 52.10 & 50.23 & 60.92 & 25.12 & \textbf{$+$22.50} & \textbf{$+$17.81} \\
            MFS3 $\dagger$ \cite{xu2023multi} & 73.42 & 69.85 & 66.44 & 62.81 & 59.78 & 56.94 & 55.04 & 53.00 & 51.07 & 60.93 & 22.35 & \textbf{$+$22.49} & \textbf{$+$15.04} \\
            PC $\dagger$ \cite{xu2023flexible} & 76.30 & 71.89 & 67.70 & 63.40 & 60.21 & 57.31 & 55.01 & 52.79 & 50.74 & 61.71 & 25.56 & \textbf{$+$21.71} & \textbf{$+$18.25} \\
            LIMIT $\dagger$ \cite{zhou2022few} & 73.81 & 72.09 & 67.87 & 63.89 & 60.70 & 57.77 & 55.67 & 53.52 & 51.23 & 61.84 & 22.58 & \textbf{$+$21.58} & \textbf{$+$15.27} \\
            FACT $\dagger$ \cite{zhou2022forward} & 74.60 & 72.09 & 67.56 & 63.52 & 61.38 & 58.36 & 56.28 & 54.24 & 52.10 & 62.24 & 22.50 & \textbf{$+$21.18} & \textbf{$+$15.19} \\
            \hline
            CLIP zero-shot & 74.45 & 72.83 & 72.11 & 70.25 & 69.71 & 69.55 & 69.52 & 68.78 & 68.04 & 70.58 & 6.40 & $+$12.84 & $-$0.91 \\
            CPE-CLIP & \textbf{87.83} & \textbf{85.86} & \textbf{84.93} & \textbf{82.85} & \textbf{82.64} & \textbf{82.42} & \textbf{82.27} & \textbf{81.44} & \textbf{80.52} & \textbf{83.42} & \textbf{7.31} & \\
            \bottomrule[1.5pt]
        \end{tabular}
    \end{center}
    \caption{CIFAR100 benchmark. $\Delta$ PD represents the improvement for PD compared to other models. $\Delta$ Avg. represents the improvement in across-session average accuracy compared to other models. $\dagger$ identifies the results taken from their respective papers, and $\ddagger$ shows the results approximated from the respective paper's figures since tabular results are unavailable.}
    \label{table:tab2}
\end{table*}

\begin{table*}
    \begin{center}
        \setlength{\tabcolsep}{3pt}
        \renewcommand{\arraystretch}{1.1}
        \begin{tabular}{lccccccccc|cc|cc}
            \toprule[1.5pt]
            \multirow{2}{*}{Method} & \multicolumn{9}{c|}{Accuracy in each session (\%)} & \multirow{2}{*}{Avg. $\uparrow$} & \multirow{2}{*}{PD $\downarrow$} & \multirow{2}{*}{$\Delta$ Avg.} & \multirow{2}{*}{$\Delta$ PD} \\
            \cline{2-10}
            & 0 & 1 & 2 & 3 & 4 & 5 & 6 & 7 & 8 & & \\
            \hline
            CEC $\dagger$ \cite{zhang2021few} & 72.00 & 66.83 & 62.97 & 59.43 & 56.70 & 53.73 & 51.19 & 49.24 & 47.63 & 57.74 & 24.37 & \textbf{$+$28.39} & \textbf{$+$16.91} \\
            CLOM $\dagger$ \cite{zou2022margin} & 73.08 & 68.09 & 64.16 & 60.41 & 57.41 & 54.29 & 51.54 & 49.37 & 48.00 & 58.52 & 25.08 & \textbf{$+$27.61} & \textbf{$+$17.62} \\
            LIMIT $\dagger$ \cite{zhou2022few} & 72.32 & 68.47 & 64.30 & 60.78 & 57.95 & 55.07 & 52.70 & 50.72 & 49.19 & 59.05 & 23.13 & \textbf{$+$27.08} & \textbf{$+$15.67} \\
            PC $\dagger$ \cite{xu2023flexible} & 73.20 & 68.35 & 64.06 & 60.85 & 58.00 & 54.98 & 52.82 & 51.17 & 50.16 & 59.28 & 23.04 & \textbf{$+$26.85} & \textbf{$+$15.58} \\
            MFS3 $\dagger$ \cite{xu2023multi} & 73.65 & 68.91 & 64.60 & 61.48 & 58.68 & 55.55 & 53.33 & 51.69 & 50.26 & 59.79 & 23.39 & \textbf{$+$26.34} & \textbf{$+$15.93} \\
            FACT $\dagger$ \cite{zhou2022forward} & 72.56 & 69.63 & 66.38 & 62.77 & 60.60 & 57.33 & 54.34 & 52.16 & 50.49 & 60.69 & 22.07 & \textbf{$+$25.44} & \textbf{$+$14.61} \\
            SPPR $\ddagger$ \cite{zhu2021self} & 80.00 & 74.00 & 68.66 & 64.33 & 61.00 & 57.33 & 54.66 & 51.66 & 49.00 & 62.29 & 31.00 & \textbf{$+$23.84} & \textbf{$+$23.54} \\
            FeSSSS $\dagger$ \cite{ahmad2022few} & 81.50 & 77.04 & 72.92 & 69.56 & 67.27 & 64.34 & 62.07 & 60.55 & 58.87 & 68.24 & 22.63 & \textbf{$+$17.89} & \textbf{$+$15.17} \\
            \hline
            CLIP zero-shot & 77.13 & 76.49 & 75.31 & 77.30 & 75.35 & 75.28 & 73.92 & 74.18 & 73.17 & 75.35 & 3.96 & $+$10.78 & $-$3.50 \\
            CPE-CLIP & \textbf{90.23} & \textbf{89.56} & \textbf{87.42} & \textbf{86.80} & \textbf{86.51} & \textbf{85.08} & \textbf{83.43} & \textbf{83.38} & \textbf{82.77} & \textbf{86.13} & \textbf{7.46} & \\
            \bottomrule[1.5pt]
        \end{tabular}
    \end{center}
    \caption{\textit{mini}ImageNet benchmark. $\Delta$ PD represents the improvement for PD compared to other models. $\Delta$ Avg. represents the improvement in across-session average accuracy compared to other models. $\dagger$ identifies the results taken from their respective papers, and $\ddagger$ shows the results approximated from the respective paper's figures since tabular results are unavailable.}
    \label{table:tab3}
\end{table*}

\begin{table*}
    \begin{center}
        \setlength{\tabcolsep}{2.3pt}
        \renewcommand{\arraystretch}{1.1}
        \begin{tabular}{lccccccccccc|cc|cc}
            \toprule[1.5pt]
            \multirow{2}{*}{Method} & \multicolumn{11}{c|}{Accuracy in each session (\%)} & \multirow{2}{*}{Avg.$\uparrow$} & \multirow{2}{*}{PD $\downarrow$} & \multirow{2}{*}{$\Delta$Avg.} & \multirow{2}{*}{$\Delta$PD} \\
            \cline{2-12}
            & 0 & 1 & 2 & 3 & 4 & 5 & 6 & 7 & 8 & 9 & 10 & & \\
            \hline
            SPPR $\dagger$ \cite{zhu2021self} & 68.68 & 61.85 & 57.43 & 52.68 & 50.19 & 46.88 & 44.65 & 43.07 & 40.17 & 39.63 & 37.33 & 49.32 & 31.35 & \textbf{$+$21.47} & \textbf{$+$14.37} \\
            PC $\dagger$ \cite{xu2023flexible} & 74.06 & 70.89 & 68.13 & 63.98 & 61.54 & 58.85 & 57.56 & 55.96 & 54.28 & 53.73 & 52.40 & 61.03 & 21.66 & \textbf{$+$9.76} & \textbf{$+$4.68} \\
            CEC $\dagger$ \cite{zhang2021few} & 75.85 & 71.94 & 68.50 & 63.50 & 62.43 & 58.27 & 57.73 & 55.81 & 54.83 & 53.52 & 52.28 & 61.34 & 23.57 & \textbf{$+$9.45} & \textbf{$+$6.59} \\
            MFS3 $\dagger$ \cite{xu2023multi} & 75.63 & 72.51 & 69.65 & 65.29 & 63.13 & 60.38 & 58.99 & 57.41 & 55.55 & 54.95 & 53.47 & 62.45 & 22.16 & \textbf{$+$8.34} & \textbf{$+$5.18} \\
            FeSSSS $\dagger$ \cite{ahmad2022few} & 79.60 & 73.46 & 70.32 & 66.38 & 63.97 & 59.63 & 58.19 & 57.56 & 55.01 & 54.31 & 52.98 & 62.85 & 26.62 & \textbf{$+$7.94} & \textbf{$+$9.64} \\
            FACT $\dagger$ \cite{zhou2022forward} & 75.90 & 73.23 & 70.84 & 66.13 & 65.56 & 62.15 & 61.74 & 59.83 & 58.41 & 57.89 & 56.94 & 64.42 & 18.96 & \textbf{$+$6.37} & \textbf{$+$1.98} \\
            LIMIT $\dagger$ \cite{zhou2022few} & 75.89 & 73.55 & 71.99 & 68.14 & 67.42 & 63.61 & 62.40 & 61.35 & 59.91 & 58.66 & 57.41 & 65.50 & 18.48 & \textbf{$+$5.29} & \textbf{$+$1.50} \\
            CLOM $\dagger$ \cite{zou2022margin} & 79.57 & 76.07 & 72.94 & 69.82 & 67.80 & 65.56 & 63.94 & 62.59 & 60.62 & 60.34 & 59.58 & 67.17 & 19.99 & \textbf{$+$3.62} & \textbf{$+$3.01} \\
            \hline
            CLIP zero-shot & 65.46 & 63.37 & 62.15 & 58.58 & 58.66 & 58.57 & 56.95 & 55.97 & 54.57 & 54.64 & 55.31 & 58.56 & 10.15 & $+$12.23 & $-$6.83 \\
            CPE-CLIP & \textbf{81.58} & \textbf{78.52} & \textbf{76.68} & \textbf{71.86} & \textbf{71.52} & \textbf{70.23} & \textbf{67.66} & \textbf{66.52} & \textbf{65.09} & \textbf{64.47} & \textbf{64.60} & \textbf{70.79} & \textbf{16.98} &  \\
            \bottomrule[1.5pt]
        \end{tabular}
    \end{center}
    \caption{CUB200 benchmark. $\Delta$ PD represents the improvement for PD compared to other models. $\Delta$ Avg. represents the improvement in across-session average accuracy compared to other models. $\dagger$ identifies the results taken from their respective papers.}
    \label{table:tab4}
\end{table*}

We have also conducted a comparison of the training time and number of learnable parameters for various models in our study. The comparison was meant to unerstand computational costs for completing the entire learning session stack. We only included models that guarantee reproducibility and have available hyperparameters. The results of this comparison are presented in Table \ref{table:tab5}. Overall, our findings indicate that CPE-CLIP significantly decreases computational costs without sacrificing performance.

\begin{table*}
    \begin{center}
        \setlength{\tabcolsep}{3pt}
        \renewcommand{\arraystretch}{1.1}
        \begin{tabular}{lcccccc}
        \toprule[1.5pt]
        & \multicolumn{2}{c}{CIFAR100} & \multicolumn{2}{c}{\textit{mini}ImageNet} & \multicolumn{2}{c}{CUB200} \\
        \cmidrule(l){2-7}
        Model & \# params. & train. time & \# params. & train. time & \# params. & train. time \\
        \hline
        CEC \cite{zhang2021few} & 295K & 0.32 & 12.2M & 1.41 & 12.3M & 0.96 \\
        FACT \cite{zhou2022forward} & 280K & 1.48 & 11.2M & 7.30 & 11.3M & 1.41 \\
        LIMIT \cite{zhou2022few} & 295K & 1.02 & 12.2M & 2.00 & 12.3M & 0.99 \\
        CLOM \cite{zou2022margin} & 350K & 0.24 & 14M & 1.56 & 18.9M & 0.35 \\
        CPE-CLIP & 400K & 0.69 & 400K & 0.65 & 400K & 0.27 \\
        \bottomrule[1.5pt]
        \end{tabular}
    \end{center}
    \caption{Training time (train. time), expressed in hours, and number of learnable parameters (\# params.). Results are obtained by simulating models from their open-source training protocol. All the simulations were performed on the same GeForce RTX 2080 Ti.}
    \label{table:tab5}
\end{table*}

\subsection{Ablation Study}

\begin{figure}[t]
    \begin{center}
        \includegraphics[width=1\linewidth]{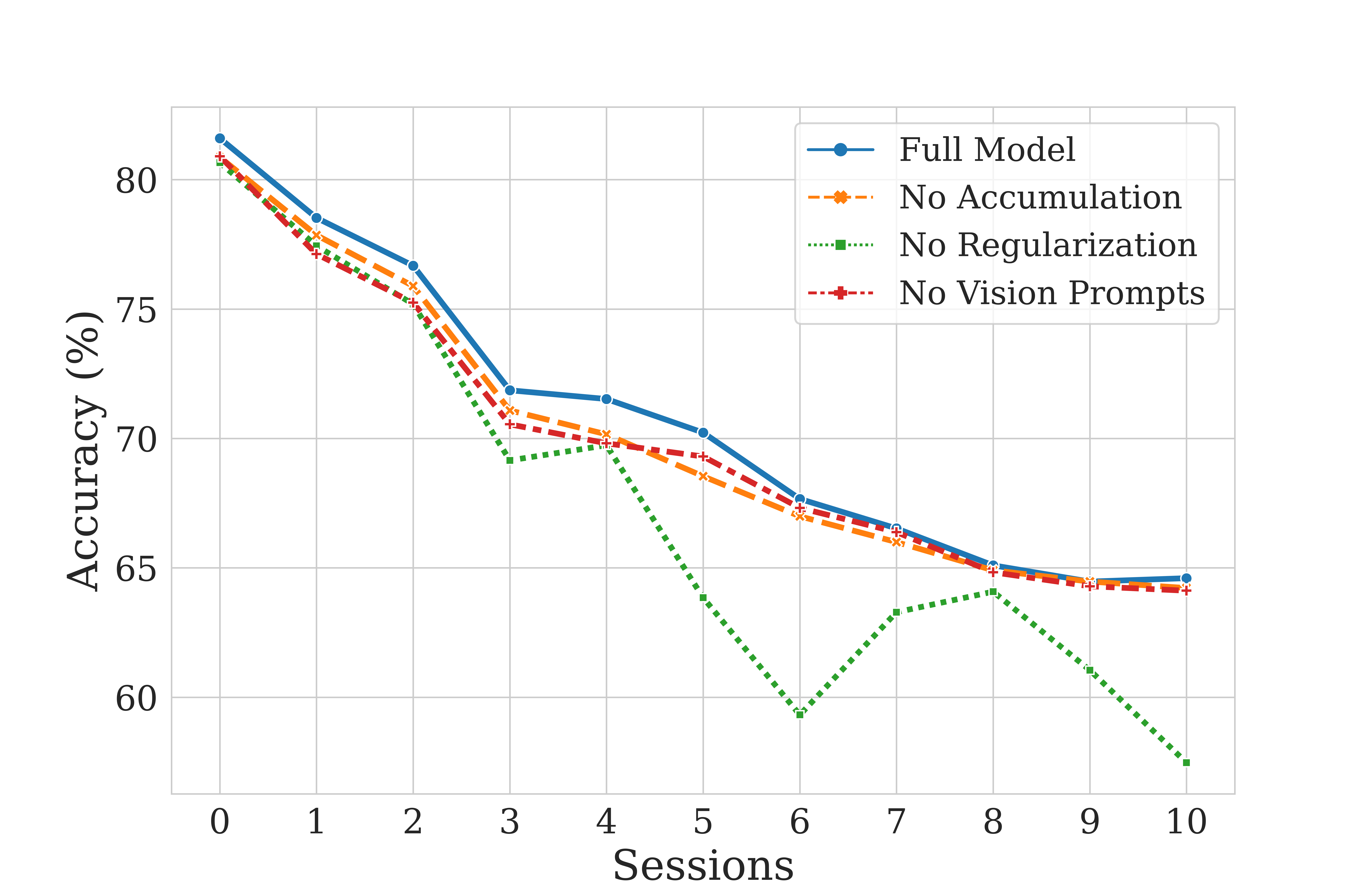}
    \end{center}
   \caption{Ablation study depicting top-1 accuracy of 5-run simulations for the main model (Full Model), and three ablated versions where accumulation is removed from the vision branch (No Accumulation), no prompts are processed by the vision branch (No Vision Prompts), and no regularization is applied (No Regularization). Session 0 refers to base class.}
    \label{fig:Fig.2}
\end{figure}

Here we analyze the importance of the relevant components in CPE-CLIP. For brevity, we only rely on the CUB200 benchmark. In particular, we focused on three ablated models. First of all, we consider the case where no accumulation strategy for prompt propagation is applied to the vision encoder. In this case, a standard replacement strategy is used, as for the language branch. Further, we focus on the contribution of projecting G-Prompt to the vision branch, by completely removing vision prompts. Finally, we consider the case where no regularization is applied so that G-Prompt updates consistently across sessions. Results are depicted in Figure \ref{fig:Fig.2}.

The comparison between the main model and its ablated versions reveals noteworthy observations. Specifically, it appears that prompt regularization is a critical element for ensuring consistent performance over time by counteracting the influence of biased distributions of few-shot training examples in problematic sessions. Conversely, the exclusion of the vision prompt system does not appear to have a marked effect on the model's susceptibility to forgetting and information loss during sessions. However, performance generally deteriorates compared to the full model throughout each session, with a noticeable decline during the initial sessions, and ultimately converges in the later ones. It is noteworthy that a minimal difference in performance is observed between the total removal of prompts from the vision branch and the utilization of a prompt propagation replacement strategy. Overall, the results confirm the effectiveness of our prompt system in achieving the best performance in a continual learning setting. To summarize, prompt regularization allows to reduce information loss and forgetting by ensuring stable training over time, and the accumulation strategy for prompt propagation in the vision encoder provides better image representation ensuring a better text-image match and higher accuracy within specific sessions.

\section{Conclusions}

Inspired by advances in few-shot image classification and parameter-efficient learning, we proposed a novel solution for solving the challenging task of Few-Shot Class Incremental Learning where a limited number of labeled data are available for each session. Our proposed CPE-CLIP effectively combined several technologies and modern ideas to conceive a multimodal few-shot continual learner that maintains high performances over time. We demonstrated that our proposed approach is capable of outperforming other approaches specifically designed for FSCIL, by relying on a small number of parameters and lower overall computational costs. CPE-CLIP introduces an accumulation strategy for prompt propagation that seems to be beneficial for enhancing image representation by ensuring the best classification accuracy. Prompt regularization ensures instead stable learning by reducing information loss over time.
\\
\textbf{Limitations.} The CPE-CLIP architecture is built upon the CLIP framework as its underlying backbone. CLIP leverages text supervision to reason about visual concepts, which serves as a primary advantage for FSCIL. However, this also poses challenges when tackling tasks that lack image category labels, that are not readily processable by the CLIP vocabulary, or that are inherently ambiguous, leading to unreliable image-text matching. Additionally, the impact of regularization on a greater number of sessions has not been explored. Although decreasing the updating rate of G-Prompt parameters as more classes are seen seems crucial to avoid over-fitting, the scaling factor for the gradient approaches zero as sessions increase. In this case, the lack of proper G-Prompt parameters update can harm generalization on novel classes when unexpected distribution shifts occur.

{\small
\bibliographystyle{ieee_fullname}
\bibliography{egbib}
}

\end{document}